\documentclass[sigconf]{acmart}
\AtBeginDocument{%
  
}
\setcopyright{none}              
\renewcommand\footnotetextcopyrightpermission[1]{} 
\acmYear{2026}
\copyrightyear{2026}
\usepackage{float}
\usepackage{needspace}
\usepackage{graphicx}
\usepackage{xcolor}
\definecolor{myblue}{RGB}{0,102,204}

\usepackage{multirow}
\usepackage{booktabs}
\usepackage{graphicx}        
\usepackage{multirow}       
\usepackage{booktabs}        
\usepackage{xcolor}           
\usepackage{amsmath}         
\usepackage{amsfonts}         
\usepackage{siunitx}

\begin{document}
\title{Conditional Polarization Guidance for Camouflaged Object Detection}

\author{Qifan Zhang}
\affiliation{%
  \institution{Dalian Maritime University}
  \city{Dalian}
  \country{China}}
\email{qifanz77@dlmu.edu.cn}

\author{Hao Wang}
\authornote{Corresponding Author}
\affiliation{%
  \institution{Dalian Maritime University}
  \city{Dalian}
  \country{China}}
\email{wh_2220233828@dlmu.edu.cn} 

\author{Xiangrong Qin}
\affiliation{%
  \institution{Dalian University of Technology}
  \city{Dalian}
  \country{China}}
\email{qxr136@mail.dlut.edu.cn}

\author{Ruijie Li}
\authornote{Project Leader}
\affiliation{%
  \institution{The Hong Kong University of Science and Technology (Guangzhou)}
  \city{Guangzhou} 
  \country{China}}
\email{rli541@connect.hkust-gz.edu.cn}

\title{Conditional Polarization Guidance for Camouflaged Object Detection}

\begin{abstract}

Camouflaged object detection (COD) aims to identify targets that are highly blended with their backgrounds. Recent works have shown that the optical characteristics of polarization cues play a significant role in improving camouflaged object detection. However, most existing polarization-based approaches depend on complex visual encoders and fusion mechanisms, leading to increased model complexity and computational overhead, while failing to fully explore how polarization can explicitly guide hierarchical RGB representation learning. To address these limitations, we propose CPGNet, an asymmetric RGB–polarization framework that introduces a conditional polarization guidance mechanism to explicitly regulate RGB feature learning for camouflaged object detection. Specifically, we design a lightweight polarization interaction module that jointly models these complementary cues and generates reliable polarization guidance in a unified manner. Unlike conventional feature fusion strategies, the proposed conditional guidance mechanism dynamically modulates RGB features using polarization priors, enabling the network to focus on subtle discrepancies between camouflaged objects and their backgrounds. Furthermore, we introduce a polarization edge-guided frequency refinement strategy that enhances high-frequency components under polarization constraints, effectively breaking camouflage patterns. Finally, we develop an iterative feedback decoder to perform coarse-to-fine feature calibration and progressively refine camouflage prediction. Extensive experiments on polarization datasets across multiple tasks, along with evaluations on non-polarization datasets, demonstrate that CPGNet consistently outperforms state-of-the-art methods.

\end{abstract}

\maketitle
\begin{figure}[t]
    \centering
    \includegraphics[width=\columnwidth]{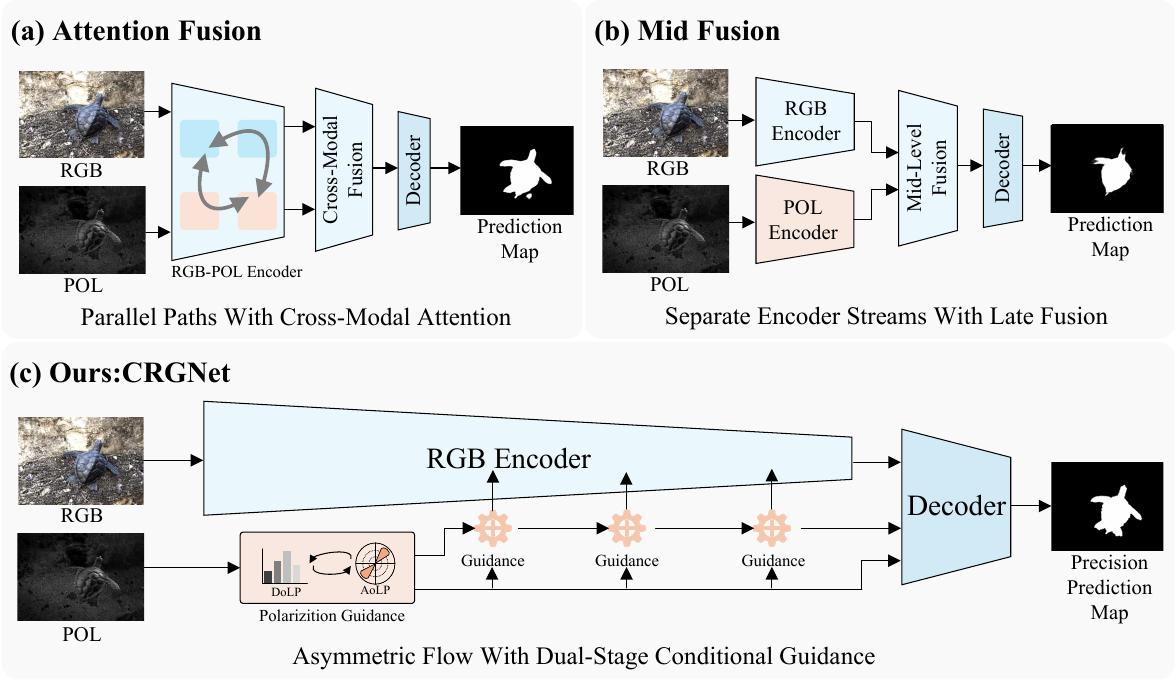}
     
    \caption{Comparison of RGB–polarization integration paradigms. (a) Attention fusion enhances RGB features using polarization-aware attention. (b) Mid-level fusion directly merges RGB and polarization features. (c) Our method treats polarization as conditional guidance for RGB representation learning, rather than explicit feature fusion.}
    \label{fig:1}
\end{figure}

\section{Introduction}
Camouflaged Object Detection (COD) aims to accurately localize and segment visually concealed objects in challenging scenes, including highly similar color, texture, and local structure with the background. Due to its broad applications in wildlife monitoring~\cite{ROY2023101919}, medical imaging~\cite{fan2020pranetparallelreverseattention}, and agriculture~\cite{wang2024depth}, COD has emerged as a critical task in computer vision.

Most existing methods rely on the RGB modality and enhance camouflage perception through strategies such as progressive localization~\cite{tankovich2023hitnethierarchicaliterativetile}, multi-scale representation learning~\cite{pang2022zoom}, boundary enhancement~\cite{Ye_2025_ICCV}, and frequency-domain modeling~\cite{sun2024frequency}.
However, the frequent occlusions and environmental interference, such as water reflections and glass refraction, make it difficult to accurately detect camouflaged objects using RGB information alone. 
Even with the advent of foundation models such as SAM~\cite{xiong2026sam2}, the performance on COD remains limited.
These limitations are especially pronounced in complex natural scenes, often leading to performance bottlenecks.

To alleviate this limitation, recent studies have introduced auxiliary modalities, such as depth~\cite{wang2023depth}, polarization~\cite{wang2024ipnet}, and infrared~\cite{11159525}, which provide complementary information beyond RGB. 
Among these modalities, polarization imaging is particularly appealing, as it captures intrinsic physical properties of light–matter interactions, revealing differences in surface materials, geometries, and reflection characteristics that are often imperceptible in RGB images. 
Such properties enable polarization to highlight structural discrepancies between camouflaged objects and their surrounding backgrounds, even when their appearance is highly similar.
In particular, the degree of linear polarization (DoLP) and the angle of linear polarization (AoLP) provide complementary cues from different aspects of polarization state, offering useful information for distinguishing subtle camouflage discrepancies.
DoLP reflects the intensity proportion of polarized light and is sensitive to material-dependent reflectance and boundary variations, while AoLP encodes the orientation of the polarization plane and captures structural and geometric information. The complementary nature of DoLP and AoLP offers valuable guidance for distinguishing subtle camouflage discrepancies and enhancing fine-grained perception in challenging scenarios.

Recent RGB-polarization COD methods have shown encouraging results, confirming the value of polarization in reducing RGB ambiguity.  For example, IPNet~\cite{wang2024ipnet} designs cross-modal modules to effectively integrate polarization and RGB cues at multiple levels, improving contrast in complex scenes.
Similarly, HIPFNet~\cite{wang2025polarization} specifically designed a polarization image enhancer that accepts Stokes parameters to optimize the unique information contained in orthogonal polarization images.
Although these methods have achieved promising performance, most existing RGB–polarization approaches, as illustrated in Fig.~\ref{fig:1}, rely on symmetric dual-stream encoding or heavy multimodal fusion, which increases model complexity while underexploring how polarization can explicitly guide hierarchical RGB representation learning.

To this end, this paper proposes a novel RGB–polarization architecture for COD, dubbed CPGNet, which leverages polarization as conditional structural guidance for RGB feature learning.
Specifically, CPGNet is composed of three coordinated components.
First, a Polarization Integration Module constructs reliable polarization guidance by exploiting the complementarity of DoLP and AoLP while suppressing less informative responses. 
Building upon this, a Polarization Guidance Flow progressively injects the derived polarization guidance into hierarchical RGB features, enabling the network to focus on subtle appearance discrepancies; meanwhile, an edge-guided frequency refinement strategy further enhances discriminative high-frequency cues to break camouflage patterns. 
Finally, an Iterative Feedback Decoder (IFD) performs coarse-to-fine prediction refinement through hierarchical feedback calibration, improving localization accuracy and boundary completeness in ambiguous camouflage regions.
Together, these components establish a unified conditional guidance framework in which polarization continuously regulates RGB representation learning rather than being directly fused as a parallel modality. Experiments on PCOD-1200 show that CPGNet performs favorably against recent RGB-only and multimodal methods. In addition, results on RGBP-Glass suggest that the proposed framework may also generalize to other RGB-polarization tasks.

Our main contributions are summarized as follows:
\begin{itemize}
    \item We introduce a new perspective for RGB-polarization camouflaged object detection, where polarization is modeled as conditional structural guidance rather than a heavily fused parallel modality.
    \item We propose CPGNet, an asymmetric RGB-polarization framework that constructs, propagates, and refines polarization guidance for hierarchical RGB representation learning.
    \item Extensive experiments on multiple RGB–polarization tasks and benchmark datasets, including  COD10K, CAMO, NC4K, and CHAMELEON, demonstrate the superiority and effectiveness of our approach. 
\end{itemize}

\section{Related Work}
\subsection{Polarization-based Vision}
Most computer vision systems rely on RGB modalities and achieve strong performance under standard conditions. However, they often struggle in challenging environments. 
To overcome the limitations of RGB-based vision, polarization, as a fundamental property of light waves, has been increasingly leveraged in both optics and computer vision tasks.
In general, polarization cameras are capable of recording four linear polarization states of light, namely $I_{0},I_{45},I_{90},I_{135}$ where $I_{x}$ represents the image captured through a linear polarizer at an angle of x. 
The polarization state of light is commonly described by the Stokes vector $S=\left[ S_0,S_1,S_2,S_3 \right] $.
\begin{align}
\mathbf{S} =
\begin{bmatrix}
S_0 \\
S_1 \\
S_2
\end{bmatrix}
=
\begin{bmatrix}
\frac{I_0 + I_{90} + I_{45} + I_{135}}{4} \\
\frac{I_0 - I_{90}}{2} \\
\frac{I_{45} - I_{135}}{2}
\end{bmatrix}
\end{align}

Based on the Stokes vector, the degree of linear polarization (DoLP) and the angle of linear polarization (AoLP) can be further defined as follows:
\begin{align}
\mathrm{DoLP}=\frac{\sqrt{S_{1}^{2}+S_{2}^{2}}}{S_0}, \quad
\mathrm{AoLP}=\frac{1}{2}\arctan \left( \frac{S_2}{S_1} \right) ,
\end{align}

DoLP and AoLP are key polarization parameters that characterize the proportion and orientation of polarized light, with DoLP measuring the ratio of linearly polarized intensity and AoLP indicating its vibration direction.  
These parameters are critical for analyzing changes in polarization states during light–object interactions and hold significant value in various optical studies and practical applications.
In the field of transparent object perception~\cite{Mei_2022_CVPR}, polarization cues provide complementary information about refractive boundaries, surface orientation, and specular behavior, thereby facilitating the detection and segmentation of transparent or glass-like objects that are highly ambiguous in RGB images. 
In the field of shape reconstruction and surface normal estimation, polarization observations encode geometric constraints induced by light–surface interactions, making them effective for recovering fine-grained surface normals and object geometry, especially for textureless or reflective surfaces. In the field of image restoration~\cite{zhou2023polarization}, polarization information can assist in recovering images degraded by scattering, absorption, and other factors, thereby improving image quality. For camouflaged object detection, polarization is particularly appealing because it is sensitive to material properties, surface orientation, and reflection characteristics, and can therefore reveal structural discrepancies that are often less evident in RGB images.

\begin{figure*}[tbp]
    \setlength{\tabcolsep}{1.0pt}
    \centering
    \small
    \begin{tabular}{c}
        \includegraphics[width=\textwidth]{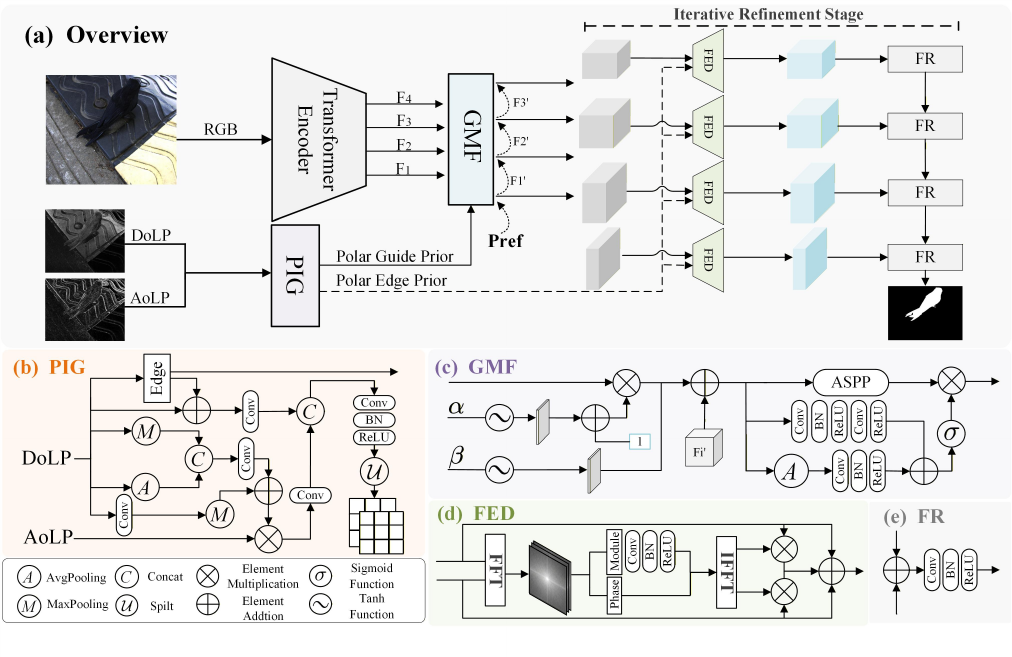}
    \end{tabular}
    \caption{(a) Overview of the proposed CPGNet. (b) Polarization Integration Module (PIM). (c) Polarization Guidance Enhancement (PGE). (d) Edge-guided Frequency Module (EFM). (e) Feature Refinement (FR). CPGNet progressively injects polarization guidance into hierarchical RGB features and refines predictions in a coarse-to-fine manner.}
    \label{fig:overview}
\end{figure*}

\subsection{Camouflaged Object Detection}
Camouflaged object detection has received increasing attention, particularly in applications such as remote sensing, wildlife monitoring, and medical imaging.
Early methods typically adopted CNNs as backbones, such as PFNet~\cite{Mei_2021_CVPR}, which introduces a positioning-and-focus strategy to progressively localize and refine camouflaged regions, and SINet~\cite{fan2020Camouflage} strengthens concealed object perception with a search-and-identification paradigm.
Transformer-based architectures have shown strong potential for global context modeling. Works like ZoomNeXt~\cite{Pang_2024}, DaCOD~\cite{wang2023depth}, PlantCamo~\cite{yang2024plantcamoplantcamouflagedetection} and CamoFormer~\cite{yin2024camoformer} adopt transformer backbones to improve representation quality.
Recently, the foundation model SAM~\cite{kirillov2023segment} has been introduced into this field and has achieved promising performance, with a series of works such as SAM-Adapter~\cite{chen2024sam2adapterevaluatingadapting}, COMPrompter~\cite{zhang2025comprompter}, and SAM-DSA~\cite{liu2025improving} further demonstrating its effectiveness.
The vision–language model CLIP has demonstrated strong transferability, prompting its adoption in COD methods such as CGNet~\cite{zhang2024cgcod} and OVCOD~\cite{OVCOS_ECCV2024}.
Despite these advances, approaches that rely solely on RGB modalities tend to degrade significantly under challenging conditions, such as low-light or overly strong illumination.
Several studies have explored auxiliary modalities for COD. PopNet~\cite{wu2023source} demonstrates the benefit of auxiliary physical priors for camouflage perception. RISNet~\cite{wang2024depth} leverages depth information for dense prediction in agricultural scenarios, while CGFINet~\cite{11159525} utilizes infrared cues to break camouflage in military settings. 
Among auxiliary modalities, polarization is particularly effective for camouflaged object detection, as it captures material, surface, and reflection cues that reveal structural differences beyond RGB. 
Most multimodal methods largely follow a symmetric dual-stream paradigm, where different modalities are encoded by parallel branches and integrated through relatively heavy cross-modal fusion. 
However, such a formulation overlooks the intrinsic asymmetry between RGB and polarization in camouflage perception: while RGB provides dominant semantic and contextual information, polarization offers physically grounded auxiliary cues that are particularly effective in revealing subtle structural discrepancies hidden in RGB images, suggesting that existing frameworks still leave room for more efficient and task-aware exploitation of polarization information.

\section{Method}
\def\w{1.0\linewidth}
\def\h{1.0in}

\subsection{Overview}
\label{sec:overview}
The overall architecture of the proposed CPGNet is illustrated in Fig.~\ref{fig:overview} (a), where polarization information is formulated as an explicit structural condition to progressively guide RGB feature learning, rather than being treated as a parallel input modality. 
First, RGB features are extracted using a Transformer-based backbone, while the lightweight Polarization Integration Module (PIM) exploits the complementary relationship between DoLP and AoLP at the input stage.
Then, each scale is processed by the Polarization Guidance Flow, which contains the Polarization Guided Enhance (PGE), ensuring that spatial features remain anchored to
structural boundaries, and the Edge-guided Frequency Module (EFM) to model fine-grained texture differences.
Finally, multi-level representations are progressively aggregated in a top-down manner, and an iterative feedback strategy is employed during decoding to capture finer-grained details.

\subsection{Polarization Integration Module}

\label{sec:pim}
DoLP $\rho_d$ and AoLP $\rho_a$ reveal object/scene materials from two different aspects. Directly using them may therefore introduce additional interference.
To address this issue, we design a Polarization Integration Module, which exploits the complementarity of AoLP and DoLP to generate robust polarization guidance for subsequent feature extraction. As the captured polarization angle is more likely random at regions with a low polarization degree, PIM filters the AoLP measurement based on the DoLP. In addition, PIM also extracts and integrates the edge information in the DoLP measurement to help distinguish objects with different materials.
This process can be expressed as:

\begin{align}
\rho_{ad} &= \rho_a \odot \Big( \phi\big([\mathcal{A}(\rho_d), \mathcal{M}(\rho_d)]\big) + \mathcal{M}\big(\phi(\rho_d)\big) \Big),\\
\gamma &= \mathrm{Edge}(\rho_d),\\
\alpha, \beta &= \chi\Big(\Phi\big([\phi(\rho_{ad}), \phi(\rho_d+\gamma)]\big)\Big),
\end{align}
where $\rho_{ad}$ denotes the polarization-guided representation, $\gamma$ denotes the DoLP-derived edge prior, and $\alpha$ and $\beta$ denote the polarization guidance parameters used for subsequent conditional modulation. $\mathcal{A}(\cdot)$ and $\mathcal{M}(\cdot)$ denote average-pooling and max-pooling respectively, $\phi(\cdot)$ denotes a convolution operation, $\Phi(\cdot)$ denotes a Conv-BN-ReLU transformation, Edge represents an edge extraction function, 
$[\cdot]$ denotes feature concatenation, $\odot$ denotes element-wise multiplication
 and $\chi(\cdot)$ denotes the channel-wise chunk operation.

\subsection{Polarization Guidance Flow}
\subsubsection{Polarization Guidance Enhance}

The RGB representations extracted from the Transformer-based backbone are denoted as $\{F_i\}_{i=1}^4$.
However, RGB features often suffer from blurred boundaries when objects exhibit low contrast against or other complex backgrounds.
The polarization cues, $\alpha, \beta$, derived from the Polarization Integration Module, provide strong structural priors that are highly sensitive to object boundaries.
Motivated by this property, we incorporate polarization-derived structural cues to guide RGB feature enhancement and improve boundary awareness. 
Compared with heavy attention-based fusion strategies, we introduce simple element-wise operations, leading to a lightweight and stable integration.
After being guided by polarization information, we perform feature enhancement and fusion across different scales in a bottom-up manner. The enhancement features are denoted as $\{E_i\}_{i=1}^4$.
This can be specified more precisely by the following equations:
\begin{align}
    F'_i = F_i \odot \big(\tau_i(\alpha) + 1\big) + \tau_i(\beta),\\
    E_i = \mathrm{ASPP}(F'_i) \odot \sigma\left(\Phi\left(\mathcal{A}(F'_i)\right) + \Phi(F'_i)\right),
\end{align}
where $F'_i$ denotes the polarization-guided RGB feature at the $i$-th stage, $\tau_i(\cdot)$ denotes a projection function, $\odot$ denotes element-wise multiplication, $\sigma(\cdot)$ denotes the sigmoid activation function, and $\Phi(\cdot)$ denotes a Conv-BN-ReLU transformation.

\subsubsection{Edge-Guided Frequency Module}

Previous studies have proved that the amplitude and phase components obtained by the Fourier transform contain more critical information for the object. 
In particular, the polarization edge structures tend to exhibit stronger responses in the amplitude spectrum.
Based on this, we leverage polarization-derived edge priors to guide frequency-domain feature enhancement. 
We apply the Fast Fourier Transform (FFT) to the RGB and edge features and perform operations on their amplitude components. After this operation, we transform again to the space domain with the Inverse Fast Fourier Transform.
This allows polarization-derived edges to provide reliable structural guidance during frequency enhancement.
This can be specified more precisely by the following equations:

\begin{align}
    \omega_r, \psi_r,\ \omega_p, \psi_p &= \mathcal{F}(E_i),\ \mathcal{F}(\gamma), \\
    \tilde{\omega}_i &= \Phi_{\omega}(\omega_i), \quad i \in \{r,p\}, \\
    \tilde{E}_i &= \mathcal{F}^{-1}\left(\tilde{\omega}_i \odot e^{j\psi_i}\right), \quad i \in \{r,p\}, \\
    \hat{E}_i &= X_i + X_i \odot \tilde{E}_i, \quad i \in \{r,p\}, \\
    \hat{E}_{\mathrm{re}} &= \hat{E}_r + \hat{E}_p,
\end{align}
where $\mathcal{F}(\cdot)$ and $\mathcal{F}^{-1}(\cdot)$ denote the Fourier transform and inverse Fourier transform, respectively; 
$\Phi_{\omega}(\cdot)$ denotes a sequence of Conv-BN-ReLU; 
$X_r = E_i$ and $X_p = \gamma$ denote the inputs of the two branches; 
$\tilde{E}_i$ is the reconstructed feature via inverse Fourier transform after amplitude refinement. 
$\hat{E}_i$ denotes the residual-enhanced feature; 
and $\hat{E}_{re}$ denotes the final frequency-enhanced representation.

\begin{figure*}[tbp]
    \centering
    \includegraphics[width=1.0\linewidth,height=9.5cm]{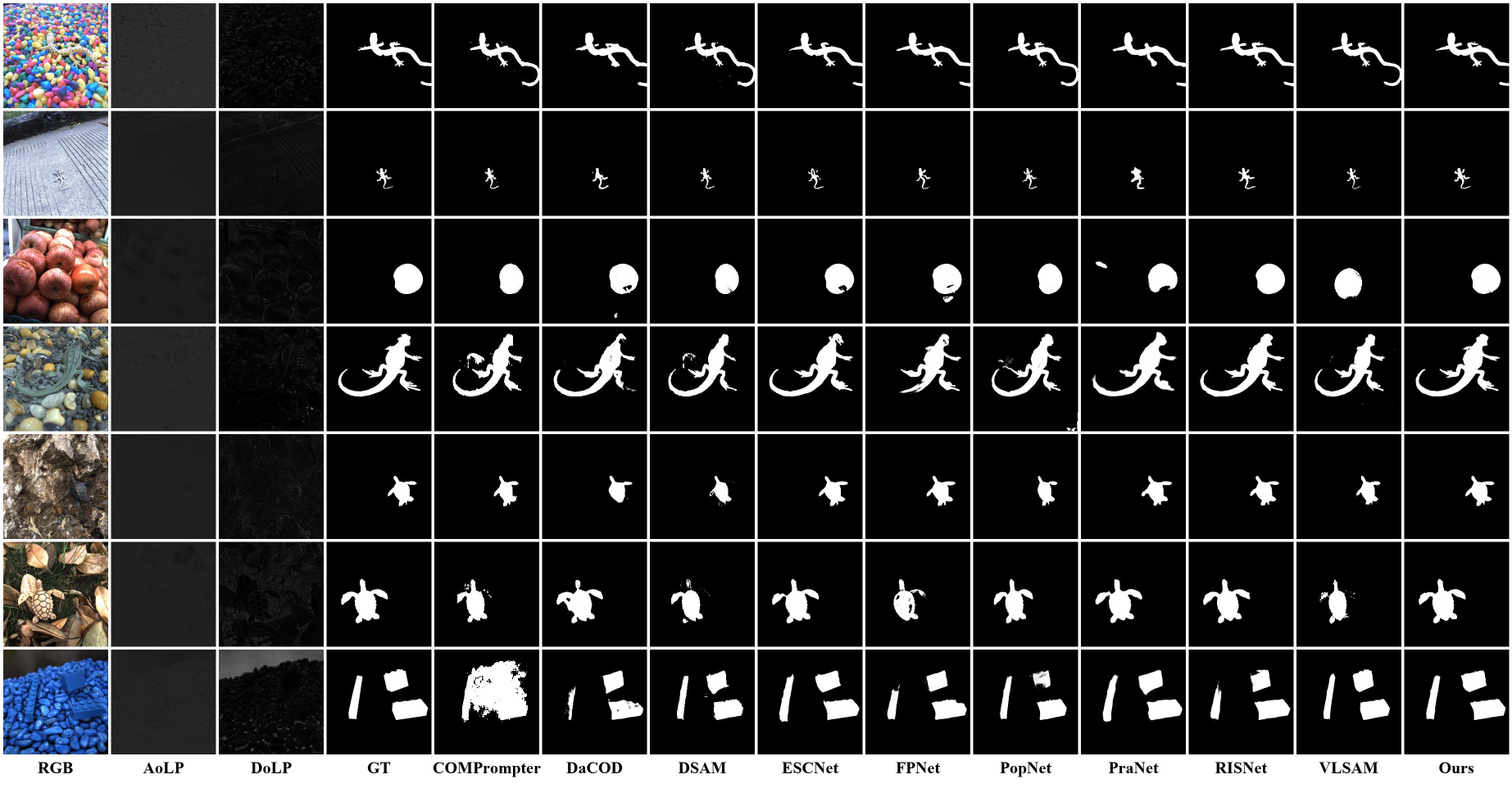}
    \caption{Qualitative comparison with SOTA methods on PCOD-1200. CPGNet produces more complete predictions and cleaner boundaries in challenging camouflage scenes.}
    \label{fig:PCOD1200}
\end{figure*}
\subsection{Iterative Feedback Decoder}
Humans often identify camouflaged objects through repeated observation, and the perception process typically begins with a coarse understanding of the scene and gradually focuses on regions of interest to refine the recognition of subtle details. This process reflects the collaborative interaction between bottom-up and top-down perceptual mechanisms. 
Inspired by this, we propose an Iterative Feedback Decoder, as illustrated in Fig.~\ref{fig:overview}(a).
The iterative process progressively refines predictions by leveraging feedback from previous stages, enabling coarse-to-fine optimization. 
This process can be formulated by the following equations:
\begin{align}
    D_i &= \phi_i\!\left(\hat{E}_{\mathrm{re}} + \operatorname{Up}(D_{i+1})\right), \quad i = 3,2,1, \\
    A^{(t-1)} &= \sigma\!\left(P^{(t-1)}\right), \\
    P^{(t)} &= f_{\mathrm{pred}}\!\left(\mathcal{D}\!\left(F'_i + \tau_i\!\left(A^{(t-1)}\right)\right)\right),
\end{align}
where $F'_i$ denotes the enhanced feature map at the $i$-th level, $D_i$ represents the decoded feature at level $i$, $\operatorname{Up}(\cdot)$ denotes the upsampling operation. 
$\mathcal{D}(\cdot)$ represents the top-down decoding process, 
and $f_{\mathrm{pred}}(\cdot)$ denotes a convolutional prediction head.$A^{(t-1)}$ denotes the feedback map generated from the prediction of the previous iteration, and $P^{(t)}$ denotes the prediction map at the $t$-th iteration.

\subsection{Loss Function}
To effectively guide each iterative stage, we apply supervision to the prediction results after each iteration of the decoder.
Specifically, we generate intermediate predictions \( P^{(i)} \) from each iteration (with the total number of iterations set to 2) and $P_4$ denotes the coarse prediction generated from the last stage.
Following~\cite{fan2020Camouflage}, we employ weighted binary cross-entropy loss \( \mathcal{L}_{\mathrm{\omega BCE}} \) and weighted intersection-over-union loss \( \mathcal{L}_{\mathrm{\omega IoU}} \) to better supervise hard-to-detect pixels, combined as $\mathcal{L}_{\mathrm{str}} = \mathcal{L}_{\mathrm{\omega IoU}} + \mathcal{L}_{\mathrm{\omega BCE}}$.
This explicit supervision ensures that both high-level semantic localization and low-level boundary refinement are jointly optimized.
Overall, our total loss function is formulated as:
\begin{equation}
\mathcal{L}_{\mathrm{total}} = \sum_{i=1}^{2} \mathcal{L}_{\mathrm{str}} \bigl( P_{(i)}, GT \bigr) + \mathcal{L}_{\mathrm{str}} \bigl( P_4, GT \bigr),
\end{equation}

\section{Experiments}

\subsection{Experimental Setup}
\noindent \textbf{Implementation Details.}
Our model is implemented in PyTorch and trained on a single RTX 4090 GPU. We employ the Adam optimizer with an initial learning rate of 1e-4, which is decayed by a factor of 10 every 50 epochs. Training is conducted for 100 epochs with a batch size of 4. All inputs, including RGB and auxiliary modalities, are resized to 704$\times$704.

\noindent \textbf{Datasets.} 
We evaluate our proposed method on the PCOD-1200 dataset~\cite{wang2024ipnet}, the largest real-world camouflage dataset with polarization information, including 970 scenes for training and 230 for testing. 
To validate the ability to perform across tasks of our proposed method, we conduct experiments on the RGBP-Glass dataset~\cite{Mei_2022_CVPR}, a widely used benchmark for transparent object detection with polarization information. 
It is currently one of the most popular public RGB-P datasets for glass object segmentation.
Furthermore, we conduct experiments on nonpolarization-based datasets for camouflage object detection, including CAMO~\cite{le2019anabranch}, COD10K~\cite{fan2020camouflaged}, NC4K~\cite{lv2021simultaneously}, and CHAMELEON~\cite{skurowski2018animal}. The training sets of CAMO and COD10K are combined for joint training, while the others are used for evaluation.

\noindent \textbf{Metrics.}
For the camouflaged object segmentation task, we adopt four widely used metrics: Mean Absolute Error ($M$), weighted $F$-measure ($F_\beta^w$), structure measure ($S_\alpha$), and enhanced alignment measure ($E_\phi$). Here, $M$ evaluates pixel-wise error, $F_\beta^w$ emphasizes precision–recall balance with spatial weighting, $S_\alpha$ measures structural similarity, and $E_\phi$ assesses both local and global alignment.
For the transparent object detection task, we employ Overall Accuracy ($OA$), Mean Absolute Error ($M$), Intersection over Union ($IoU$), Balanced Error Rate ($BER$), and $F$-measure ($F_\beta$). Specifically, $OA$ reflects overall accuracy, $IoU$ evaluates region overlap, $BER$ balances foreground–background errors, and $F_\beta$ captures the precision–recall trade-off.
\begin{table}[tbp]
\centering
\footnotesize
\scalebox{1.0}{
\begin{tabular}{p{2cm}p{1.5cm}p{0.6cm}p{0.6cm}p{0.6cm}p{0.6cm}}
\hline
Methods &Pub.'Year& $S_\alpha\uparrow$ &$E_\phi\uparrow$ & $M\downarrow$ &$F_{\beta}^w\uparrow$ \\
\hline
\multicolumn{6}{c}{\textbf{\textit{RGB-only Methods}}} \\
\hline
PFNet~\cite{Mei_2021_CVPR}&CVPR'21 &0.876 &0.946&0.013&0.788\\
SINet-V2~\cite{fan2020Camouflage}&TPAMI'22& 0.882&0.941&0.013& -\\
HQSAM*~\cite{sam_hq}&NeurIPS'23&0.936&0.973&0.008&0.909\\
FEDER~\cite{He2023Camouflaged}&CVPR’23&0.892 &0.946 &0.011&0.843\\
FSEL~\cite{sun2024frequency}&ECCV'24 &0.934&0.978&0.006&-\\
PRNet~\cite{10379651}&TCSVT'24&0.933&0.978 &0.007 &0.894\\
HGINet~\cite{yao2024hierarchical}&TIP'24&0.930&0.972&0.006&-\\
ZoomNeXt~\cite{Pang_2024}&TPAMI'24&0.945&0.974&0.006&0.909\\
ESCNet~\cite{Ye_2025_ICCV}&ICCV'25&0.925&0.969&0.007&0.881\\
UTNet~\cite{11175541}&TMM'25&0.769&0.891&0.025&0.648\\
PlantCamo~\cite{yang2024plantcamoplantcamouflagedetection}&AIR'25&0.942&\textbf{0.981}&0.006&0.913\\
MM-SAM*~\cite{ren2025multi}&ICCV'25&0.892&0.925&0.013&0.801\\
SAM2-UNet*~\cite{xiong2026sam2}& Vis.Intell'26 &0.923 & 0.966  &0.009 &0.871\\
\hline
\multicolumn{6}{c}{\textbf{\textit{Multimodal Methods}}} \\
\hline
PGSNet~\cite{Mei_2022_CVPR} &CVPR'22 & 0.916&0.965&0.010& -\\
DaCOD~\cite{wang2023depth}& MM'23& 0.896& 0.955 & 0.011& 0.823 \\
PopNet~\cite{wu2023source}&ICCV'23 & 0.898 & 0.956 & 0.010 & 0.833 \\
RISNet~\cite{wang2024depth}& CVPR'24& 0.922 & 0.967 &0.008& 0.881\\
IPNet~\cite{wang2024ipnet}&EAAI'24&0.922 &0.970 &0.008 & -\\
DSAM*~\cite{yu2024exploring}&MM'24&0.907&0.952&0.013&0.839\\
COMPrompter*~\cite{zhang2025comprompter}&SCIS'25&0.886&0.929&0.019&0.795\\
HIPFNet~\cite{wang2025polarization}&EAAI'25&0.944&0.980&\textbf{0.005}&-\\
\hline
Ours &  &\textbf{0.951}&\textbf{0.981}&\textbf{0.005}&\textbf{0.923}\\
\hline
\end{tabular}
}
\caption{Quantitative comparisons with SOTAs on PCOD-1200. The best results are highlighted in \textbf{bold}. * represents the SAM-based methods.}
\label{tab:polardata}
\end{table}

\begin{table}[tbp]
\centering
\footnotesize
\setlength{\tabcolsep}{7pt} 
\scalebox{0.95}{
\begin{tabular}{l|ccccc}
\hline
Methods & $OA$ (\%) $\uparrow$ & $M$ $\downarrow$ & $IoU$ (\%) $\uparrow$ & $BER$ (\%) $\downarrow$ & $F_\beta$ (\%) $\uparrow$ \\
\hline
MirrorNet$_{19}$~\cite{Yang_2019_ICCV} & 89.44 & 0.121 & 79.22 & 11.30 & 87.02 \\
SINetv2$_{22}$~\cite{fan2020Camouflage} & 88.78 & 0.117 & 79.85 & 11.05 & 86.73  \\
PGSNet$_{22}$~\cite{Mei_2022_CVPR} & 90.98 & 0.091 & 81.8 & 9.63 & 89.38 \\
PolarNet$_{23}$~\cite{WANG2023106} & 91.15 & 0.099 & 83.11 & 9.66 & 89.84 \\
IPNet$_{24}$~\cite{wang2024ipnet} & 91.08 & 0.098 & 81.83 & 8.97 & 90.15 \\
RISNet$_{24}$~\cite{wang2024depth} & 84.54 & 0.158 & 73.70 & 15.21 & 82.10  \\
ESCNet$_{25}$~\cite{Ye_2025_ICCV} & 91.67 & 0.084 & 84.57 & 8.17 & 89.77  \\
FuseISP$_{25}$~\cite{FAN2025113429} & \textbf{92.39} & 0.088 & 83.83 & 8.47 &\textbf{ 91.05} \\
\hline
Ours & 92.05 & \textbf{0.082} & \textbf{85.09} & \textbf{7.84} & 90.45 \\
\hline
\end{tabular}
}
\caption{Quantitative comparisons with multimodal SOTAs on the \textbf{RGBP-Glass} dataset. The best results are highlighted in \textbf{bold}.}
\label{tab:glass_multimodal}
\end{table}

\subsection{Qualitative and Quantitative Evaluation}

\noindent\textbf{Quantitative Analysis.}
We quantitatively compare our method with 20 state-of-the-art (SOTA) methods on the PCOD-1200 dataset, including 12 RGB-based methods and 8 multimodal segmentation methods. As shown in Tab.~\ref{tab:polardata}, our method achieves the best performance across all evaluation metrics, with scores of 0.951 in $S_{\alpha}$, 0.923 in $F_{\beta}^w$, 0.981 in $E_{\phi}$, and 0.005 in $M$. Our method consistently outperforms all SAM-based methods, demonstrating its superiority in capturing challenging object characteristics. Compared with the second-best model HIPFNet~\cite{wang2025polarization}, our method achieves a significant improvement of 0.007 in $S_{\alpha}$. The strong $S_{\alpha}$ result further indicates that the proposed conditional polarization guidance effectively preserves informative structural cues for camouflaged object detection.
\begin{figure}[tbp]
    \centering
    
 \includegraphics[width=1.0\linewidth,height=5.5cm]{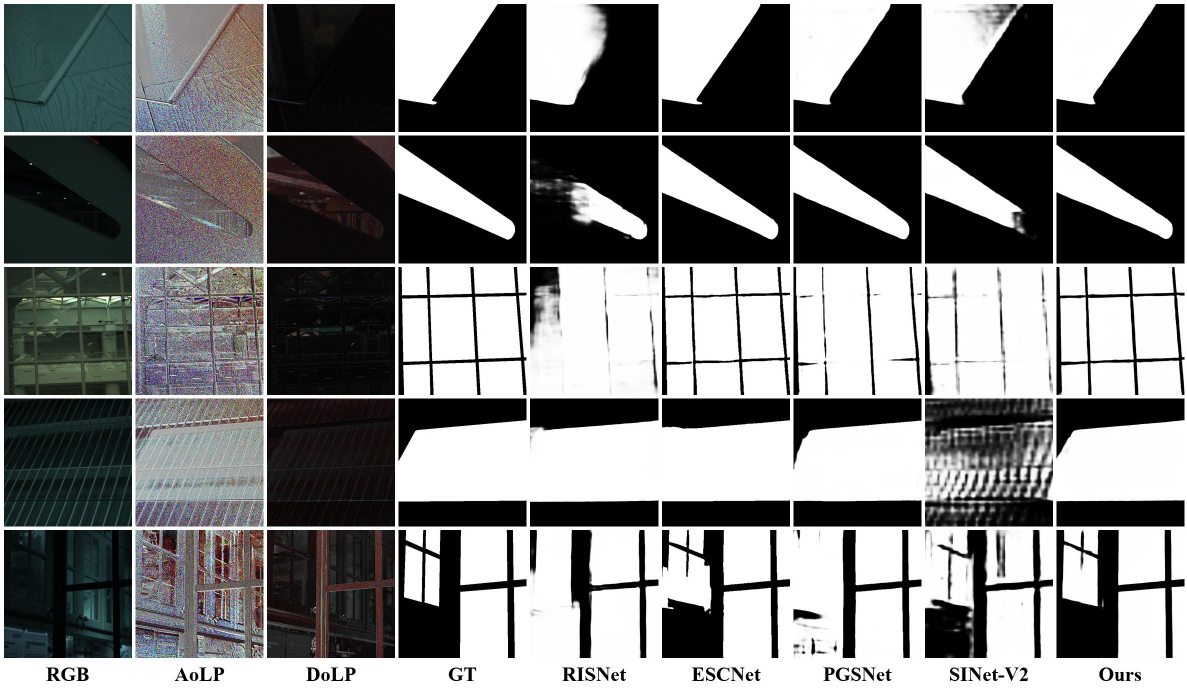}
    \caption{Qualitative comparison of our CPGNet against SOTA approaches on the RGBP-Glass dataset.}
    \label{fig:RGBPGlass}
\end{figure}

\noindent\textbf{Qualitative Analysis.}
Fig.~\ref{fig:PCOD1200} presents a visual comparison between our proposed method and other advanced methods. We evaluate challenging camouflaged images featuring different object types under extreme scenes involving multiple targets and small objects. Our approach produces cleaner and more complete predictions, especially in cluttered backgrounds with strong texture interference. These results indicate that our model effectively captures rich object details. AoLP and DoLP provide effective auxiliary cues due to their sensitivity to material properties, thereby improving the discriminability of camouflaged targets. A closer inspection of the visual results further reveals that the advantage of our method becomes more evident in scenes with complex textures, weak object boundaries, and severe appearance ambiguity. In these cases, several competing methods either miss object parts or produce scattered false positives in the background. By contrast, our method achieves more reliable localization in these challenging cases. This observation suggests that the proposed conditional guidance mechanism helps the network better exploit the complementary structural and material cues provided by polarization.
\begin{figure*}[tbp]
    \centering
    \includegraphics[width=\linewidth,height=8cm]{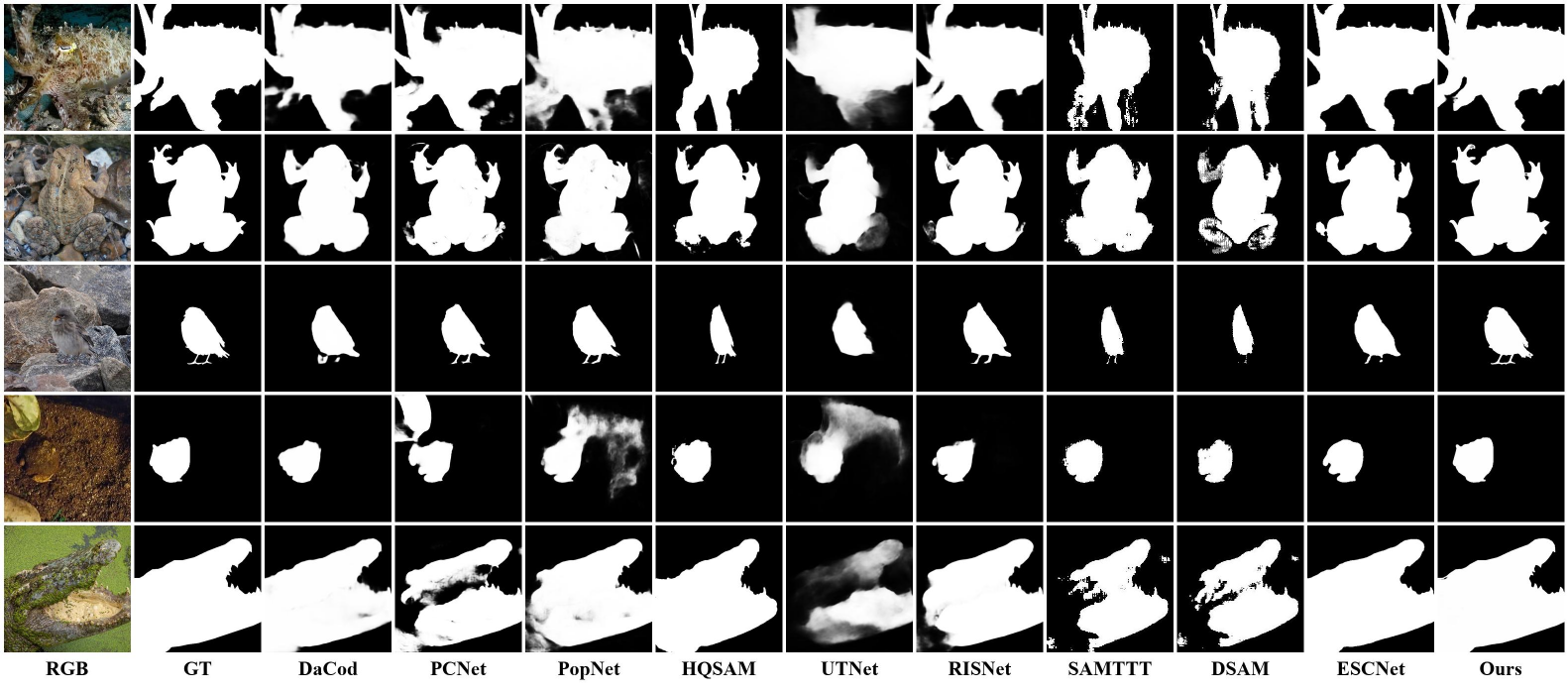}
    \caption{Qualitative comparison of our CPGNet against SOTA approaches on the RGB-Depth datasets.}
    \label{fig:RGBD}
\end{figure*}
\begin{table*}[h]
    \centering
    \footnotesize
    \setlength{\tabcolsep}{3pt}
    \renewcommand{\arraystretch}{1.05}
    \resizebox{\textwidth}{!}{%
        \begin{tabular}{p{1.6cm}|p{1.2cm}|cccc|cccc|cccc|cccc}
            \hline
            \hline
            \multirow{2}{*}{Methods} & \multirow{2}{*}{Params(M)↓} 
            & \multicolumn{4}{c|}{CAMO} 
            & \multicolumn{4}{c|}{COD10K} 
            & \multicolumn{4}{c|}{NC4K} 
            & \multicolumn{4}{c}{CHAMELEON} \\
            \cline{3-18}
            & 
            & $S_{\alpha}\uparrow$ & $F_{\beta}^w\uparrow$ & $E_{\phi}\uparrow$ & $M\downarrow$
            & $S_{\alpha}\uparrow$ & $F_{\beta}^w\uparrow$ & $E_{\phi}\uparrow$ & $M\downarrow$
            & $S_{\alpha}\uparrow$ & $F_{\beta}^w\uparrow$ & $E_{\phi}\uparrow$ & $M\downarrow$
            & $S_{\alpha}\uparrow$ & $F_{\beta}^w\uparrow$ & $E_{\phi}\uparrow$ & $M\downarrow$ \\
            \hline
            PFNet~\cite{Mei_2021_CVPR}&48.55M&0.782&0.695&0.845&0.085&0.800&0.660&0.880&0.040&0.829&0.745&0.891&0.053&0.882&0.810&0.927&0.053\\
            SPSN~\cite{leespsn}&34.04M&0.773&0.782&0.829&0.084&0.789&0.727&0.854&0.042&0.852&0.852&0.908&0.043&-&-&-&-\\
            PopNet~\cite{wu2023source}&198.9M&0.808&0.744&0.859&0.077&0.851&0.757&0.910&0.028&0.861&0.808&0.910&0.042&0.910&0.893&-&0.022\\
            PlantCamo~\cite{yang2024plantcamoplantcamouflagedetection}&27.66 M& 0.819& 0.757& 0.875& 0.076&0.856 &0.767 & 0.911& 0.029&0.853 &0.789 & 0.899& 0.047&0.899&0.853&0.941&0.030\\
            UTNet~\cite{11175541}& 10.02M&0.677&0.518&0.733&0.123&0.699&0.485&0.773&0.068&0.743&0.603&0.798&0.086&0.818&0.709&0.885&0.054\\
            FEDER~\cite{He2023Camouflaged}&42.09M&0.802&0.738&0.867&0.071&0.822&0.751&0.900&0.032&0.847&0.789&0.907&0.044&-&-&-&-\\
            \hline
            Ours  & 25.14M& \textbf{0.852} & \textbf{0.793} & \textbf{0.899} & \textbf{0.059} & \textbf{0.870} & \textbf{0.786} & \textbf{0.921} & \textbf{0.026} & \textbf{0.877} & 0.820& \textbf{0.916} & \textbf{0.039} & \textbf{0.930} & 0.892 & \textbf{0.962} & \textbf{0.019} \\
            \hline
        \end{tabular}%
    }
    \caption{Comparison of SOTA methods on multiple RGB-Depth datasets in terms of $S_{\alpha}$, $F_{\beta}^{w}$, $E_{\phi}$, and $M$. The best results are highlighted in \textbf{bold}.}
    \label{tab:RGBDdata}
\end{table*}

\noindent\textbf{Comparisons on Transparent Object Detection Datasets.}
To further assess the generality of the proposed framework, we additionally evaluate it on the RGBP-Glass dataset, a related RGB-polarization glass segmentation benchmark. 
We quantitatively compare our method with 8 methods on the RGBP-glass dataset.
As shown in Tab.~\ref{tab:glass_multimodal}, our method achieves the best performance on $M$, $IoU$, and $BER$, with scores of 0.082, 85.09, and 7.84, respectively, while also obtaining competitive results on other metrics.
This suggests that the proposed conditional polarization guidance can generalize well to other RGB-polarization dense prediction tasks.
As shown in Fig.~\ref{fig:RGBPGlass}, our method achieves accurate segmentation of transparent objects under challenging conditions, such as complex backgrounds or varying illumination.

\noindent\textbf{Comparisons on Non-polarization Datasets.}
To further assess the generalization capability of our framework, we evaluate it on widely used COD benchmarks, including CAMO, COD10K, and NC4K. 
As these datasets do not provide polarization measurements, we approximate polarization cues using surrogate signals, where Sobel-based RGB gradients and estimated depth maps are adopted to mimic DoLP and AoLP, respectively. 
As shown in Tab.~\ref{tab:RGBDdata}, our method performs favorably against other competitive COD methods on these datasets.
As shown in Fig.~\ref{fig:RGBD}, our method demonstrates superior performance in challenging scenarios, particularly exhibiting more accurate boundary prediction for camouflaged objects.
While our method is primarily designed for polarization-driven perception and is not specifically optimized for RGB-only settings, it still achieves competitive performance with a relatively low number of parameters, demonstrating the robustness and transferability of the proposed polarization-guided design.

\subsection{Ablation Study}
We conduct ablation studies on the PCOD-1200 dataset to validate the effectiveness of the proposed CPGNet. Since the proposed framework consists of several coordinated designs, we mainly adopt a removal-based ablation protocol, where each variant is constructed by removing or simplifying one component from the full model.

\begin{table}[htbp]
\centering
\small   
\setlength{\tabcolsep}{6.5pt}      
\renewcommand{\arraystretch}{1.} 
\begin{tabular}{l|lcccc}
\hline
\multirow{1}{*}{\#} & \multirow{1}{*}{Networks} & $S_\alpha \uparrow$ & $E_\phi \uparrow$ & $M \downarrow$ & $F_{\beta}^w \uparrow$ \\
\hline
\textit{A} &RGB  only           & 0.942 & 0.976 & 0.006 & 0.909 \\
\textit{B} &RGB+AoLP             & 0.945 & 0.979 & 0.006 & 0.915 \\
\textit{C} &RGB+DoLP             & 0.947 & 0.980 & 0.005& 0.917 \\
\hline
\textit{D}& w/o PGE             & 0.944 & 0.979 & 0.006 & 0.910 \\
\textit{E} & PGE   w/o $\alpha,\beta$  & 0.947 & 0.980 & 0.006 & 0.918 \\
\textit{F}&  PGE   w/o Enhance         & 0.943 & 0.976 & 0.005 & 0.910 \\
\hline
\textit{G} &w/o EFM             & 0.943 & 0.974 & 0.006 & 0.908 \\
\textit{H} &EFM   w/o Polar    & 0.944 & 0.977 & 0.006 & 0.909 \\
\hline
\textit{I} &w/o Iter           & 0.939 & 0.973 & 0.007 & 0.894 \\
\hline
Ours  &          & \textbf{0.951} & \textbf{0.981} & \textbf{0.005} & \textbf{0.923} \\
\hline
\end{tabular}
\caption{Ablation study on the PCOD-1200 dataset.}
\label{tab:abli}
\end{table}
\begin{figure*}[h]
    \centering
    
 \includegraphics[width=\linewidth,height=6cm]{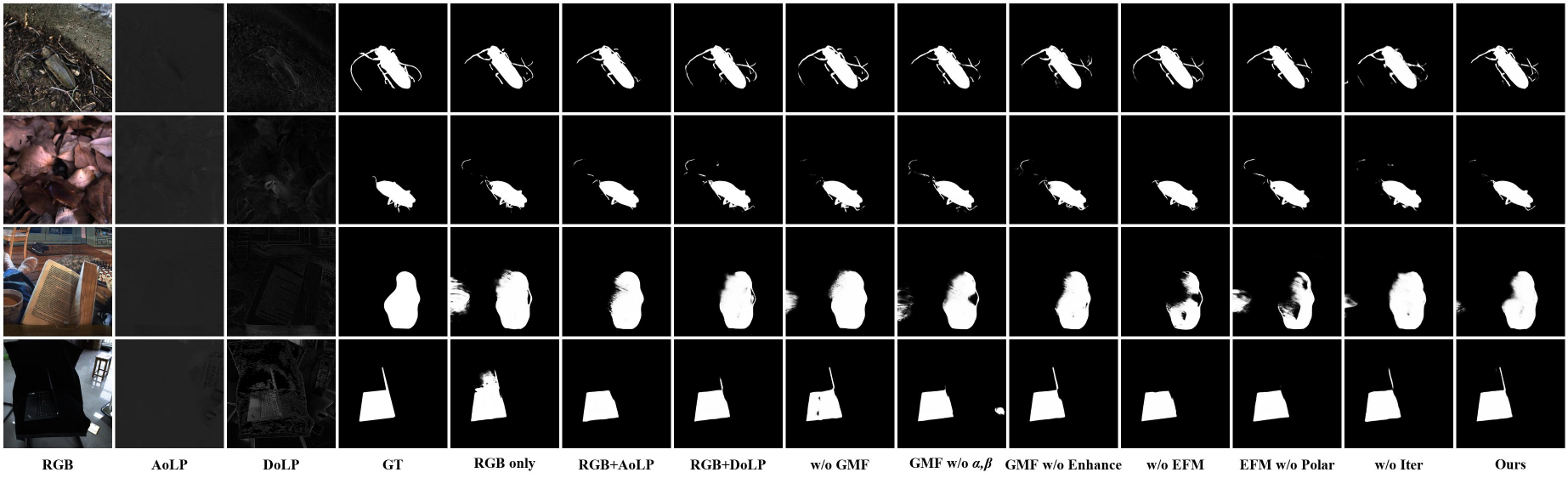}
   \caption{Qualitative ablation results on PCOD-1200. Each component improves prediction completeness and boundary quality, and the full model yields the best visual results.}
\label{fig:ablation_vis}
\end{figure*}
\noindent\textbf{Impact of Polarization Cues.}
As shown in Tab.~\ref{tab:abli}, incorporating polarization cues consistently improves the performance over the RGB-only baseline. Compared with baseline (A), introducing AoLP (B) leads to gains in structure-related metrics, including $S_{\alpha}$ and $E_{\phi}$, indicating that AoLP provides useful directional and structural information for camouflage perception. Introducing DoLP alone (C) further improves the results, suggesting that DoLP also provides effective complementary cues for distinguishing camouflaged targets. These observations indicate that both AoLP and DoLP are beneficial, while capturing different aspects of polarization information. However, each cue alone can only provide partial polarization evidence. This motivates our design to jointly model them for constructing more reliable polarization guidance, and supports our design philosophy of treating polarization as structural guidance rather than direct feature fusion.

\noindent\textbf{Effectiveness of PGF and Its Components.}
We further analyze the effectiveness of the proposed Polarization Guidance Flow (PGF), together with its two key components, namely Polarization Guidance Enhance (PGE) and Edge-guided Frequency Module (EFM). 
Removing the entire PGF (D) leads to clear performance degradation, indicating that explicit polarization-guided feature regulation is essential for enhancing hierarchical RGB representations. 
When the affine guidance parameters $\alpha$ and $\beta$ in PGE are discarded (E), the performance also drops, showing that these learned guidance parameters play an important role in transferring polarization priors into RGB feature learning. 
Further removing the enhancement design in PGE (F) causes additional degradation, which suggests that simple guidance alone is insufficient without subsequent feature reinforcement.

The EFM also contributes positively to the final performance. 
Removing EFM (G) degrades the overall results, validating the effectiveness of frequency-domain enhancement for capturing subtle high-frequency discrepancies in camouflaged regions. 
Moreover, replacing the polarization-aware EFM with a variant without polarization priors (H) further weakens the results, demonstrating that the benefit does not come merely from frequency processing itself, but from the polarization-constrained refinement mechanism.

\noindent\textbf{Effectiveness of the Iterative Feedback Decoder.}
Finally, removing the Iterative Feedback Decoder (IFD) (I) results in the most noticeable performance drop, especially in $F_{\beta}^{w}$ and $S_{\alpha}$, indicating that coarse-to-fine feedback refinement is crucial for recovering ambiguous structures and improving boundary completeness. As shown in Fig.~\ref{fig:ablation_vis}, the qualitative results further support our findings.

\noindent\textbf{Overall Discussion.} Overall, all ablation variants perform worse than the full model, which confirms that the superiority of CPGNet comes from the coordinated effect of PIM, PGF, and IFD, where PGF further benefits from the joint contribution of PGE and EFM. This also suggests that effective camouflage perception requires not only reliable polarization guidance, but also its progressive propagation and refinement throughout the network.

\begin{figure}[t]
   \centering
    
   \includegraphics[width=\linewidth]{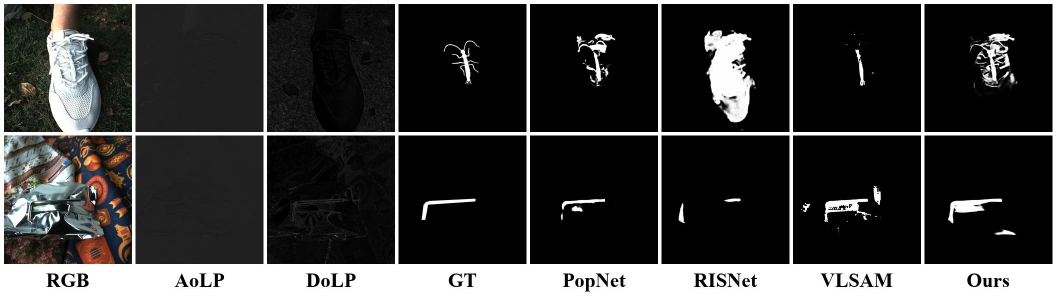}
   \caption{Failure cases caused by degraded polarization quality. When DoLP/AoLP measurements are weak, noisy, or unstable, the derived guidance may become unreliable, leading to inaccurate localization and incomplete segmentation.}
   \label{fig:failure}
\end{figure}
\section{Limitation}

The proposed CPGNet still has several limitations. Its performance depends on the quality of polarization measurements, and unstable DoLP/AoLP responses in weak-polarization regions may affect the reliability of the derived guidance cues.
Fig.~\ref{fig:failure} shows that RGB lacks discriminative features, while both polarization and depth modalities offer limited complementary cues, making it challenging for many methods to detect camouflaged targets.

\section{Conclusion}

This work presents CPGNet, a framework that introduces an asymmetric conditional polarization guidance paradigm for camouflaged object detection. By rethinking polarization not as a parallel modality for heavy fusion but as structural guidance for RGB feature learning, our approach enables more efficient and task-aware cross-modal interaction.
Our framework consists of three key components: 
a polarization integration module that leverages DoLP–AoLP complementarity for stable structural guidance, 
a polarization guidance flow that progressively modulates RGB features with polarization priors and enhances subtle discrepancies via edge-guided frequency refinement, 
and an iterative feedback decoder that performs coarse-to-fine refinement to improve ambiguous region recovery and boundary completeness.
Extensive experiments on RGB–polarization tasks demonstrate the robustness and effectiveness of our method, while additional evaluations on widely used COD benchmarks further validate its strong generalization and transferability.
In the future, we will explore the integration of polarization with foundation models or multimodal large models, which may further unlock its potential in complex visual understanding tasks.

\bibliographystyle{ACM-Reference-Format}
\bibliography{ref}
\appendix

\section{Additional Implementation Details}
To complement the main paper, we briefly provide additional implementation details of CPGNet.
The overall framework follows the design described in the main text, where polarization is used as conditional guidance for hierarchical RGB representation learning rather than being treated as a heavily fused parallel stream.

In our implementation, RGB features are extracted by a Transformer-based backbone and then projected into a unified feature space for subsequent multi-stage processing.
For polarization modeling, AoLP and DoLP are first integrated by a lightweight polarization interaction branch.
Specifically, AoLP is reweighted according to DoLP-aware responses, while DoLP is further enhanced with an edge prior.
The resulting polarization representation is then used to generate stage-wise guidance for hierarchical RGB modulation.

For conditional guidance, we adopt lightweight projection heads to produce modulation parameters for different RGB stages.
These guidance parameters are injected into multi-level RGB features through simple affine-style feature transformation, which keeps the interaction efficient and stable.
After guidance injection, each stage is further processed by a lightweight enhancement block for feature refinement.

To strengthen subtle structural discrepancies, we additionally employ a frequency enhancement design.
In practice, the refined feature is normalized and transformed into the frequency domain, where its spectral magnitude is adaptively adjusted and then projected back to the spatial domain.
Meanwhile, a polarization-derived edge prior is projected and fused into the enhanced feature representation, providing additional structural constraints during refinement.

Finally, the decoder adopts a top-down refinement manner together with iterative feedback.
A coarse prediction is first generated from high-level representations, and the refined feature from the current decoding stage is further fed back to improve the next prediction round.
Unless otherwise specified, the training protocol follows the settings described in the main paper.
\begin{figure}[!t]
    \centering
    \includegraphics[width=\linewidth]{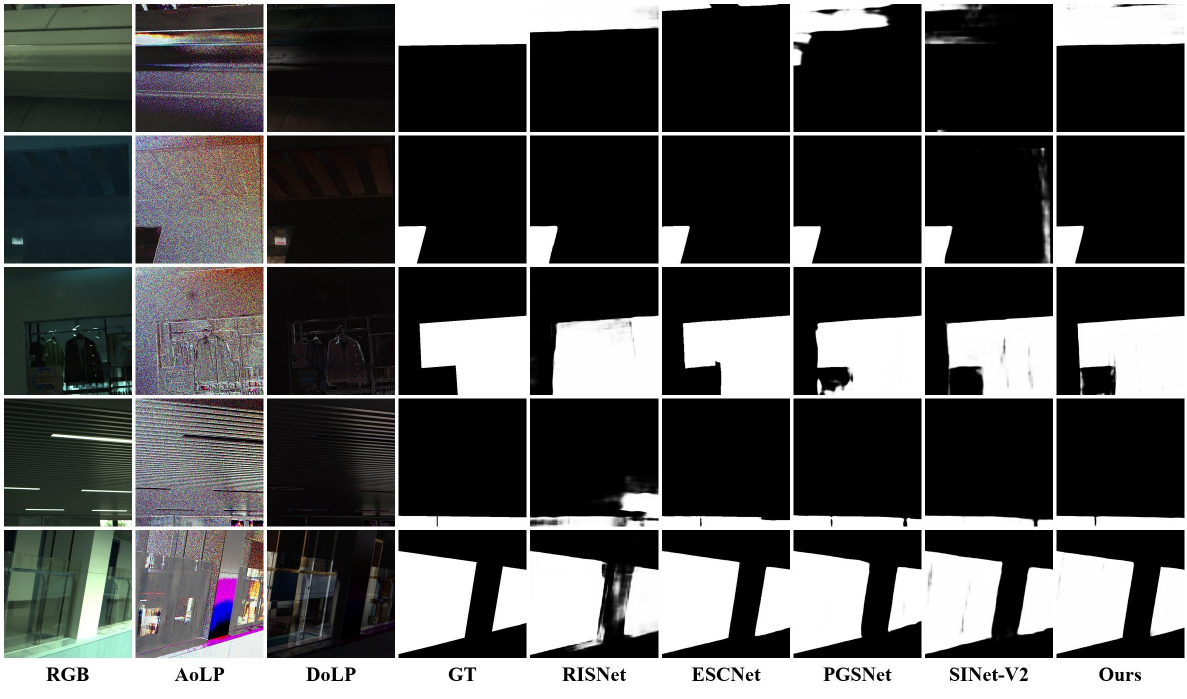}
    \caption{Additional qualitative comparisons on RGBP-Glass dataset.}
    \label{fig:supp_glass}
\end{figure}
\section{Additional Experimental Results}
We provide supplementary experimental observations to further support the effectiveness of the proposed framework.
Consistent with the ablation results in the main paper, the complete model achieves the best overall performance, indicating that the superiority of CPGNet comes from the coordinated effect of reliable polarization integration, progressive guidance propagation, frequency-aware enhancement, and iterative decoding.

More specifically, the additional observations support three main points.
First, jointly exploiting AoLP and DoLP is more effective than using either cue alone, suggesting that the two polarization components provide complementary guidance for camouflage perception.
Second, removing the polarization-guided enhancement or the frequency refinement design leads to degraded predictions, especially around ambiguous boundaries and texture-confused regions.
Third, iterative feedback contributes to more complete object structures and better boundary continuity, showing the importance of coarse-to-fine refinement for challenging camouflage scenes.

\begin{figure*}[!htbp]
    \centering
    \includegraphics[width=\linewidth]{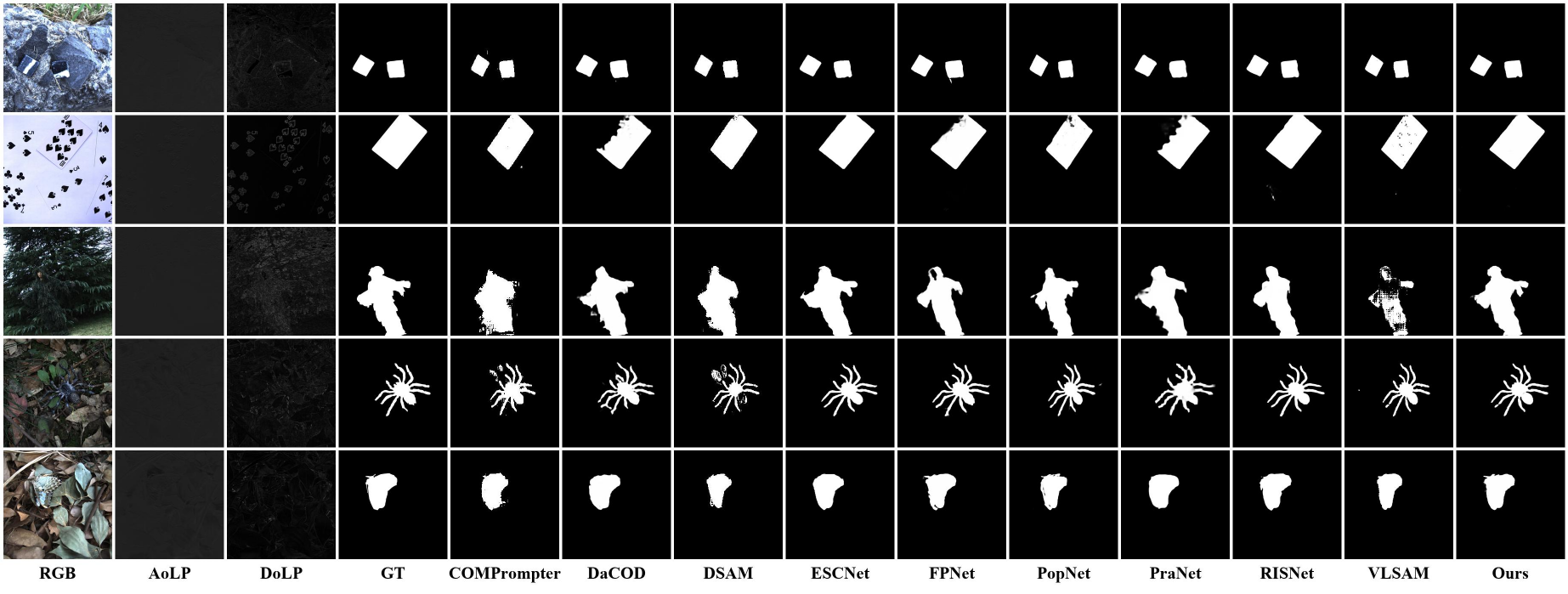}
    \caption{Additional qualitative comparisons on PCOD-1200.}
    \label{fig:supp_pcod}
\end{figure*}

\begin{figure*}[!htbp]
    \centering
    \includegraphics[width=\textwidth]{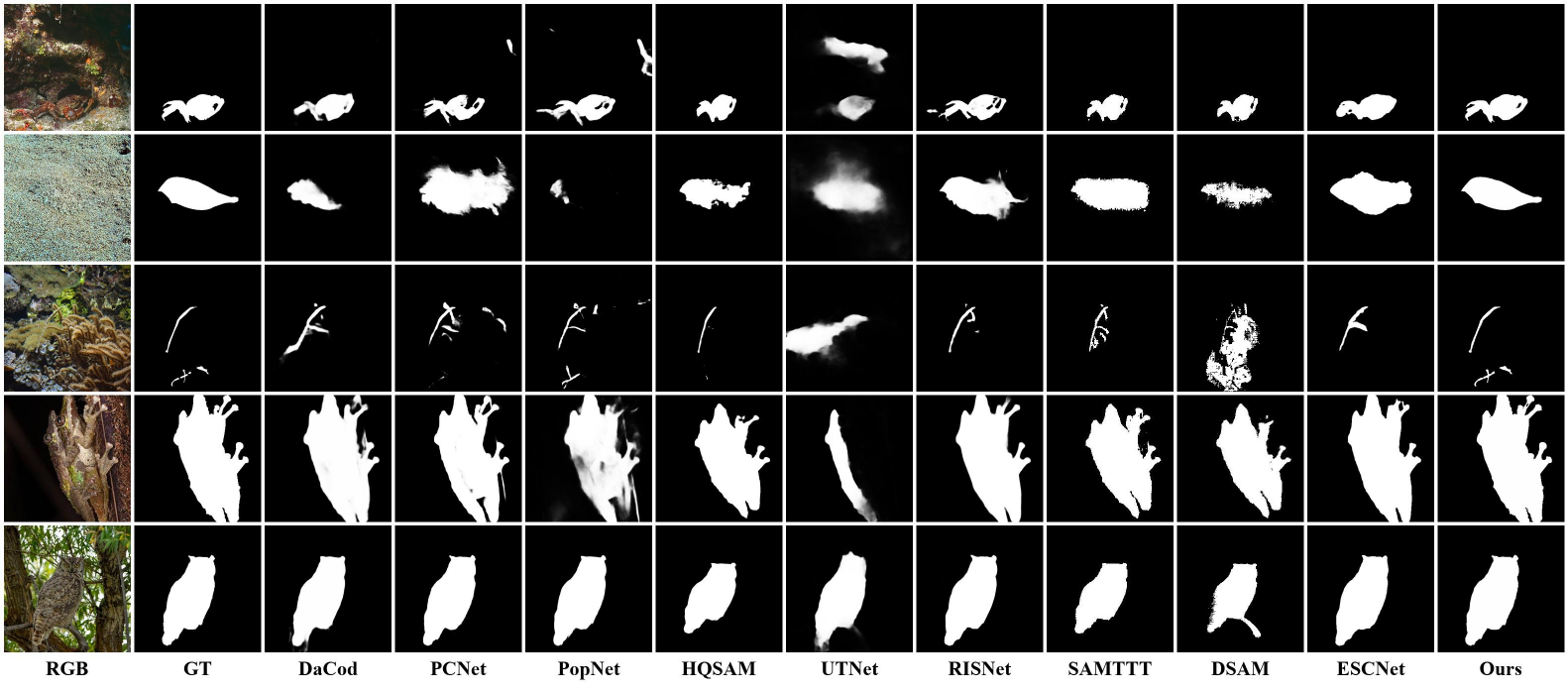}
    \caption{Additional qualitative comparisons on RGB-Depth datasets.}
    \label{fig:supp_depth}
\end{figure*}

\section{Additional Qualitative Comparisons}
We include additional qualitative comparisons on PCOD-1200, RGBP-Glass, and non-polarization COD benchmarks. 
As shown in Fig.~\ref{fig:supp_pcod}, Fig.~\ref{fig:supp_glass}, and Fig.~\ref{fig:supp_depth}, these examples consistently show that the proposed method produces more complete foreground regions and cleaner object boundaries under challenging conditions, such as weak structural contrast, cluttered backgrounds, and severe texture interference.

Compared with competing methods, our predictions are generally more stable in ambiguous regions and less likely to introduce scattered false positives.
This tendency is particularly evident when the object boundary is weak or when the appearance contrast between foreground and background is extremely low.
The qualitative results in Fig.~\ref{fig:supp_pcod} further demonstrate the effectiveness of our method on polarization-based camouflage scenes, while Fig.~\ref{fig:supp_glass} and Fig.~\ref{fig:supp_depth} show that the proposed framework also generalizes well to transparent object detection and non-polarization COD benchmarks, respectively.
These additional visual results further support the effectiveness of the proposed conditional polarization guidance mechanism.

\clearpage
\end{document}